\documentclass{article}
\usepackage{ijcai19}
\usepackage{bm}
\usepackage{amsthm}
\usepackage{amsmath}%
\usepackage{amsfonts}
\usepackage{balance}
\usepackage{times}
\usepackage{soul}
\usepackage{url}
\usepackage[hidelinks]{hyperref}
\usepackage[utf8]{inputenc}
\usepackage[small]{caption}
\usepackage{graphicx}
\usepackage{amsmath}
\usepackage{booktabs}
\usepackage{algorithm}
\usepackage{algorithmic}
\usepackage[tight,footnotesize]{subfigure}
\usepackage{fancyhdr}
\title{Efficient Relaxed Gradient Support Pursuit for Sparsity\\ Constrained Non-convex Optimization}

\author{
Fanhua Shang$^\dag$\and
Bingkun Wei$^\dag$\and
Hongying Liu$^\dag$\and
Yuanyuan Liu$^\dag$\And
Jiacheng Zhuo$^\ddag$\\
\affiliations
$^\dag$Key Laboratory of Intelligent Perception and Image Understanding of Ministry of Education,\\
School of Artificial Intelligence, Xidian University, China\\
$^\ddag$Department of Computer Science, The University of Texas at Austin
\emails
$\{$fhshang, hyliu, yyliu$\}$@xidian.edu.cn, \;bkwei028@gmail.com,\; jzhuo@cs.utexas.edu
}

\begin{document}

\maketitle

\begin{abstract}
Large-scale non-convex sparsity-constrained problems have recently gained extensive attention. Most existing deterministic optimization methods (e.g., GraSP) are not suitable for large-scale and high-dimensional problems, and thus stochastic optimization methods with hard thresholding (e.g.,  SVRGHT) become more attractive. Inspired by GraSP, this paper proposes a new general relaxed gradient support pursuit (RGraSP) framework, in which the sub-algorithm only requires to satisfy a slack descent condition. We also design two specific semi-stochastic gradient hard thresholding algorithms. In particular, our algorithms have much less hard thresholding operations than SVRGHT, and their average per-iteration cost is much lower (i.e., $O(d)$ vs.\ $O(d\log(d))$ for SVRGHT), which leads to faster convergence. Our experimental results on both synthetic and real-world datasets show that our algorithms are superior to the state-of-the-art gradient hard thresholding methods.
\end{abstract}

\section{Introduction}
Massive high-dimensional data are common nowadays and impose new challenges to algorithms for sparse learning. For high-dimensional data analysis, it is important to exploit the low intrinsic structure and dimensionality of the data, such as sparsity and low-rank structures, which is often attained by imposing certain structural assumptions on the parameter of the underlying model. In recent years, the $\ell_{1}$-norm regularized models, such as Lasso \cite{10.2307/2346178} and $\ell_{1}$-norm regularized logistic regression \cite{vandegeer2008,DBLP:conf/nips/NegahbanRWY09}, were proposed to pursue computational tractability by using the $\ell_{1}$-norm as a surrogate to the $\ell_{0}$-norm. In spite of computational advantages, the $\ell_{1}$-norm models have some limits \cite{candes2008enhancing} and attain worse empirical performance than $\ell_{0}$-norm models \cite{fan2001variable}. Thus, it is necessary and challenging to solve the $\ell_{0}$-norm constrained problem directly. In this paper, we focus on the following sparsity-constrained optimization problem,
\begin{equation}\label{equ1}
\min_{\textbf{x}\in\mathbb{R}^d}\;\mathcal{F}(\textbf{x})=\frac{1}{n}\sum_{i=1}^n f_i(\textbf{x}),\;\, \mathrm{s.t.},\,\|\textbf{x}\|_0 \leq s,
\end{equation}
where $\mathcal{F}(\textbf{x})$ is the sum of a finite number of smooth convex component functions $f_i(\cdot)$, and $\|\textbf{x}\|_0 \!\leq\!s$ means that the number of nonzero entries in $\textbf{x}$ is no more than $s$, and $s$ is used to control the sparsity level of the model parameter. The problem (\ref{equ1}) arises in machine learning, statistics and related areas, e.g., the sparsity-constrained linear regression problem:
\begin{equation}\label{equ2}
\min_{\textbf{x}\in\mathbb{R}^d}\;\frac{1}{2n}\!\sum_{i=1}^n(y_i-\textbf{w}_i^T\textbf{x})^2,\;\:\mathrm{s.t.},\, \|\textbf{x}\|_0 \leq s,
\end{equation}
where $\textbf{x}\!\in\!\mathbb{R}^d$ is the unknown model parameter, $\textbf{y}\!=\![y_1,\ldots,y_n]^T\!\in\!\mathbb{R}^n$ is the response vector, and $\textbf{W}\!=\![\textbf{w}_1,\ldots,\textbf{w}_n]\!\in\!\mathbb{R}^{d\times n}$ is a design matrix.

Due to the non-convexity of the sparsity constraint, Problem (\ref{equ1}) is NP-hard. In order to obtain an approximate solution to Problem (\ref{equ1}), a large family of greedy algorithms
\cite{DBLP:journals/tsp/MallatZ93,NeedellT10,DBLP:journals/tit/TroppG07,DBLP:journals/siamnum/Foucart11,DBLP:journals/siamjo/Shalev-ShwartzSZ10,DBLP:journals/jmlr/BahmaniRB13,DBLP:journals/tit/Zhang11,Tian:2016:FBG:3020948.3021024} have been proposed. Besides them, there has been much progress towards gradient hard thresholding methods such as fast gradient hard thresholding pursuit (FG-HT) \cite{yuan2014gradient}, iterative hard thresholding (IHT) \cite{blumensath2009iterative}. However, all the algorithms are based on deterministic optimization such as gradient descent. In each iteration, the gradient descent algorithms require the computation of a full gradient over very large $n$ component functions, which is expensive in solving large-scale problems (i.e., $O(nd)$).

To address this issue, several stochastic optimization algorithms have been proposed. For example, Nguyen \emph{et al.}~\shortcite{DBLP:journals/corr/NguyenNW14} proposed a stochastic gradient hard thresholding algorithm (SG-HT), and Li \emph{et al.}~\shortcite{DBLP:journals/corr/LiZALH16} proposed a stochastic variance reduced gradient hard thresholding algorithm (SVRGHT), which is based on the stochastic variance reduction gradient (SVRG, also called semi-stochastic gradient in \cite{konevcny2017semi}) method \cite{DBLP:conf/nips/Johnson013}. With the help of variance reduction technique, SVRGHT can converge stably and efficiently, and also obtain a better estimation accuracy than SG-HT \cite{shen:HTSVRG}. Besides these methods, there are still several stochastic first- or second-order nonconvex optimization algorithms, such as ASBCDHT \cite{DBLP:conf/uai/ChenG16}, HSG-HT \cite{DBLP:conf/nips/ZhouYF18a}, FNHTP \cite{DBLP:conf/kdd/ChenG17} and SL-BFGS \cite{DBLP:conf/ijcai/GaoH18a}. However, most algorithms mentioned above need one hard thresholding operation in each inner-iteration, which is time-consuming for high-dimensional data (i.e., $O(d\log(d))$ in general and we can also improve the time complexity using a max-heap strategy). On the other hand, a hard thresholding operation used in each inner-iteration breaks the information of current solution, which may require more gradient descent steps to reach the same accuracy. It should be emphasized that in this paper, we do not consider the coordinate-descent type algorithms \cite{Nesterov2012cd,beck2013sparsity,DBLP:conf/uai/ChenG16}. As a result, a new algorithm that needs less hard thresholding operations and yet remains fast stochastic updates becomes more attractive for large-scale and high-dimensional problems.

Inspired by the gradient support pursuit (GraSP) \cite{DBLP:journals/jmlr/BahmaniRB13} and compressive sampling matching
pursuit (CoSaMP) \cite{NeedellT10} algorithms, this paper proposes a new relaxed gradient support pursuit (RGraSP) framework to solve large-scale sparsity-constrained problems. In each iteration of our framework, we first find the most relevant support set, minimize slackly over the support set by an algorithm, which only requires to satisfy a certain descent condition, and then perform a hard thresholding operator on the updated parameter. For minimizing objective functions, we introduce two efficient semi-stochastic gradient algorithms into our RGraSP framework and propose a stochastic variance reduced gradient support pursuit (SVRGSP) algorithm and its fast version (SVRGSP+). Moreover, benefiting from significantly less hard thresholding operations than SVRGHT, the average per-iteration computational cost in our algorithms is much lower (i.e., $O(d)$ for our algorithms vs.\ $O(d\log(d))$ for the algorithms mentioned above such as SVRGHT), which leads to faster convergence. Experimental results on synthetic and real-world datasets verify the superiority of our algorithms against the state-of-the-art methods.

\begin{table}[t]
\label{tab_sim1}
\caption{Comparison of the per-iteration complexities of some sparsity-constrained optimization methods.}
\renewcommand\arraystretch{1.5}
\setlength{\tabcolsep}{2.16pt}
\begin{tabular}{lc}
\hline
Algorithms  & Per-iteration Complexity\\
\hline
FG-HT and IHT (deterministic)   & $O(nd)$\\
SG-HT and SVRGHT (stochastic) & $O(d\log(d))$ \\
SVRGSP and SVRGSP+ (Ours)       & $O(d)$\\
\hline
\end{tabular}
\end{table}

\section{Notations}
Throughout this paper, we use $\textbf{W}\!=\![\textbf{w}_1,\textbf{w}_2,\ldots,\textbf{w}_n]\!\in \!\mathbb{R}^{d\times n}$ to denote the design matrix,  $\textbf{y}\!=\![y_1,y_2,\ldots,y_n]^T\!\in\! \mathbb{R}^n$ to denote the response vector, and  $\textbf{x}\!=\![x_1,x_2,\ldots,x_d]^T\!\in\! \mathbb{R}^d$ to denote the model parameter. $\textbf{x}^*$ is the optimal solution of Problem (\ref{equ1}) and $s^*$ is the optimal sparsity level which satisfies $\|\textbf{x}^*\|_0\leq s^*$. For the parameter $\textbf{x}\!\in\! \mathbb{R}^d$, $\|\textbf{x}\|_0$ is the number of nonzero entries in the vector $\textbf{x}$, $\|\textbf{x}\|_1\!=\!\sum_{i=1}^d\! |x_i|$ is the $\ell_1$-norm, and $\|\textbf{x}\|_2\!=\!\sqrt{\sum_{i=1}^d x_i^2}$ is the $\ell_2$-norm. $\textup{supp}$($\textbf{x}$) denotes the index set of nonzero entries of $\textbf{x}$, and $\textup{supp}(\textbf{x},s)$ is the index set of the largest $s$ entries of $\textbf{x}$ in terms of magnitude. $\mathcal{T}^c$ denotes the complement set of $\mathcal{T}$, \,$\textbf{x}|_\mathcal{T}$ is a vector that equals $\textbf{x}$ except for coordinates in $\mathcal{T}^c$ where it is zero, and $|\mathcal{T}|$ denotes the cardinality of $\mathcal{T}$. In addition, $\nabla{\mathcal{F}(\textbf{x})}$ denotes the gradient of the objective function $\mathcal{F}(\cdot)$ at $\textbf{x}$, and $I$ is an identity matrix.

\section{Relaxed Gradient Support Pursuit}
In this section, we propose an efficient relaxed gradient support pursuit framework for sparsity-constrained non-convex optimization problems. Moreover, we also present the details of two specific stochastic variance reduced gradient support pursuit algorithms (called SVRGSP and SVRGSP+).

\subsection{Our Gradient Support Pursuit Framework}
Most of existing gradient support pursuit algorithms use deterministic optimization methods to minimize various sparsity-constrained problems (e.g., Problem (1)). However, there are several stochastic algorithms such as SVRGHT \cite{DBLP:journals/corr/LiZALH16}, which can be used to attain better performance \cite{DBLP:journals/corr/LiZALH16,shen:HTSVRG,DBLP:conf/ijcai/GaoH18a}.  Inspired by the well-known GraSP \cite{DBLP:journals/jmlr/BahmaniRB13} and CoSaMP \cite{NeedellT10}, this paper proposes a Relaxed Gradient Support Pursuit (RGraSP) framework to quickly find an approximate solution to Problem (1). As pointed out in \cite{DBLP:journals/jmlr/BahmaniRB13}, CoSaMP can be viewed as a special case of GraSP, when the squared error $\mathcal{F}(\textbf{x})\!=\!\frac{1}{2n}\!\sum_{i=1}^n(y_i-\textbf{w}_i^T\textbf{x})^2$ is the cost function.

\renewcommand{\baselinestretch}{1.39}
\begin{algorithm}[t]
\caption{\!: Relaxed Gradient Support Pursuit Framework}
\label{alg1}
\renewcommand{\algorithmicrequire}{\textbf{Input:}}
\renewcommand{\algorithmicensure}{\textbf{Initialize:}}
\renewcommand{\algorithmicoutput}{\textbf{Output:}}
\begin{algorithmic}[1]
\REQUIRE $\mathcal{F}(\cdot)$, $s$, and the number of iterations, $T$.\\
\ENSURE $\hat{\textbf{x}}^{0}$.\\
\FOR {$t=1,2,\cdots,T$}
\STATE {Compute local gradient: $\textbf{g}=\nabla{\mathcal{F}(\hat{\textbf{x}}^{t-1})}$;}
\STATE {Identify directions: $\mathcal{Z}=\textup{supp}(\textbf{g},2s)$;}
\STATE {Merge supports: $\mathcal{T}=\mathcal{Z}\cup\textup{supp}(\hat{\textbf{x}}^{t-1})$;}
\STATE {Minimize slackly: find $\textbf{b}$, s.t., $\|\textbf{b}\!-\!\hat{\textbf{b}}\|_2\!\leq\! c_1\|\hat{\textbf{x}}^{t-1}\!-\!\hat{\textbf{b}}\|_2$, where $\hat{\textbf{b}}$ is an optimal solution to Problem (\ref{sub}), and set $\textbf{b}|_{\mathcal{T}^c}=\textbf{0}$;}
\STATE {Perform hard thresholding over $\mathcal{T}$: $\hat{\textbf{x}}^{t}=\mathcal{H}_s(\textbf{b})$;}
\ENDFOR
\OUTPUT {$\hat{\textbf{x}}^{T}$}.
\end{algorithmic}
\end{algorithm}
\renewcommand{\baselinestretch}{1.0}

The main difference between GraSP \cite{DBLP:journals/jmlr/BahmaniRB13} and our RGraSP framework is that the former needs to yield the exact solution $\hat{\textbf{b}}$ to the following problem: $$\hat{\textbf{b}}=\arg\min_{\textbf{x}\in\mathbb{R}^d}\;\mathcal{F}(\textbf{x}),\;\;\textup{s.t.,}\;\;{\textbf{x}\mid_{\mathcal{T}^c}=\textbf{0}},$$
while the latter only requires a solver (e.g., Algorithm 2 below) for an approximation solution $\textbf{b}$ to the above problem. RGraSP is used to solve sparsity-constrained non-convex optimization problems by allowing users to pick a specially designed algorithm according to the properties of $\mathcal{F}(\cdot)$. In other words, we can choose different solvers to solve the sub-problem in Step 5 of Algorithm 1, as long as the algorithm satisfies a certain descent condition. In this sense, both GraSP \cite{DBLP:journals/jmlr/BahmaniRB13} and CoSaMP \cite{NeedellT10} can be viewed as a special case of RGraSP, when $\textbf{b}\!=\!\hat{\textbf{b}}$. Our general RGraSP framework is outlined in Algorithm \ref{alg1}.

At each iteration of Algorithm \ref{alg1}, we first compute the gradient of $\mathcal{F}(\cdot)$ at the current estimate, i.e., $\textbf{g}\!=\!\nabla\! \mathcal{F}(\hat{\textbf{x}}^{t-1})$. Then we choose $2s$ coordinates of $\textbf{g}$ that have the largest magnitude as the direction, in which pursuing the minimization will be most effective, and denote their indices by $\mathcal{Z}$, where $s$ is the sparsity constant. Merging the support of the current estimate with the $2s$ coordinates mentioned above, we can obtain the combined support, which is a set of at most $3s$ indices, i.e., $\mathcal{T}\!=\!\mathcal{Z}\!\cup\! \textup{supp}(\hat{\textbf{x}}^{t-1})$. Over $\mathcal{T}$, we compute an estimate $\textbf{b}$ as the approximate solution to the sub-problem,
\begin{gather}\label{sub}
\min_{\textbf{x}\in\mathbb{R}^d}\;\mathcal{F}(\textbf{x})=\frac{1}{n}\sum_{i=1}^n f_i(\textbf{x})~~\textup{s.t.},~{\textbf{x}\mid_{\mathcal{T}^c}=\textbf{0}}.
\end{gather}
The parameter $\hat{\textbf{x}}^{t}$ is then updated using the hard thresholding operator, which keeps the largest $s$ terms of the intermediate estimate $\textbf{b}$. This step makes $\hat{\textbf{x}}^{t}$ as the best $s$-term approximation of the estimate $\textbf{b}$. The hard thresholding operator is defined as follows:
$$
[\mathcal{H}_s(\textbf{x})]_i=
\left\{ \begin{aligned}
&x_i, &\textup{if}~i\in \textup{supp}(\textbf{x},s),\\
&0,  &\textup{otherwise}.
\end{aligned} \right.
$$
Essentially, the hard thresholding operator $\mathcal{H}_s(\textbf{x})$ keeps the largest $s$ (in magnitude) entries and sets the other entries equal to zero.

In Algorithm \ref{alg1}, we only require that the solver for solving the sub-problem (\ref{sub}) has a performance guarantee as:
$$\|\textbf{b}-\hat{\textbf{b}}\|_2\leq c_1\|\hat{\textbf{x}}^{t-1}-\hat{\textbf{b}}\|_2,$$
where $\hat{\textbf{b}}$ is an optimal solution to the sub-problem (\ref{sub}), and $0\!<\!c_1\!<\!1$ is an error tolerance, which implies that our solver proposed below has to achieve a certain accuracy for our RGraSP framework. In fact, our solver can approach to a given accuracy after sufficient iterations, as suggested by \cite{zhu:Katyusha}. Although there are quite a number of solvers that we can use, such as the first-order solvers \cite{DBLP:journals/jmlr/BahmaniRB13,liu:saga,zhou:saga} and the second-order solvers \cite{andrew2007bfgs,DBLP:conf/kdd/ChenG17,DBLP:conf/ijcai/GaoH18a}, we observe that the semi-stochastic gradient solver outperforms other solvers in most cases as in \cite{liu:ssgd}. In the following, we will present two efficient semi-stochastic gradient algorithms as our solver in Algorithm \ref{alg1}.

\begin{figure*}[t]
\centering
\subfigure[$n=2500,\,d=5000,\,s^*\!=250$]{\includegraphics[width=0.505\columnwidth]{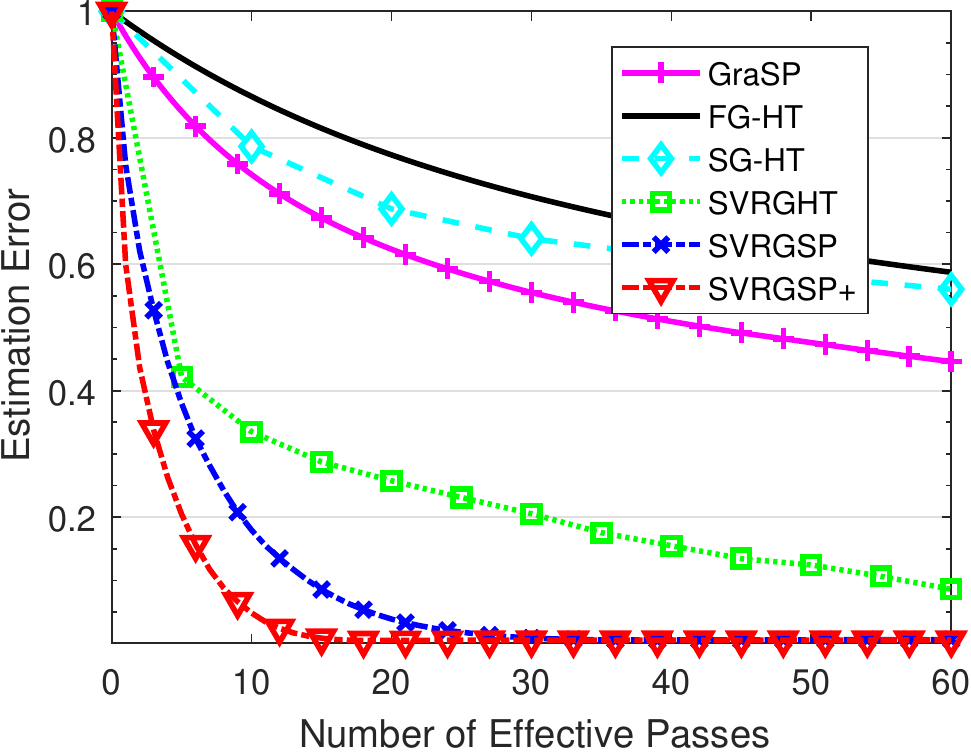}\,\includegraphics[width=0.505\columnwidth]{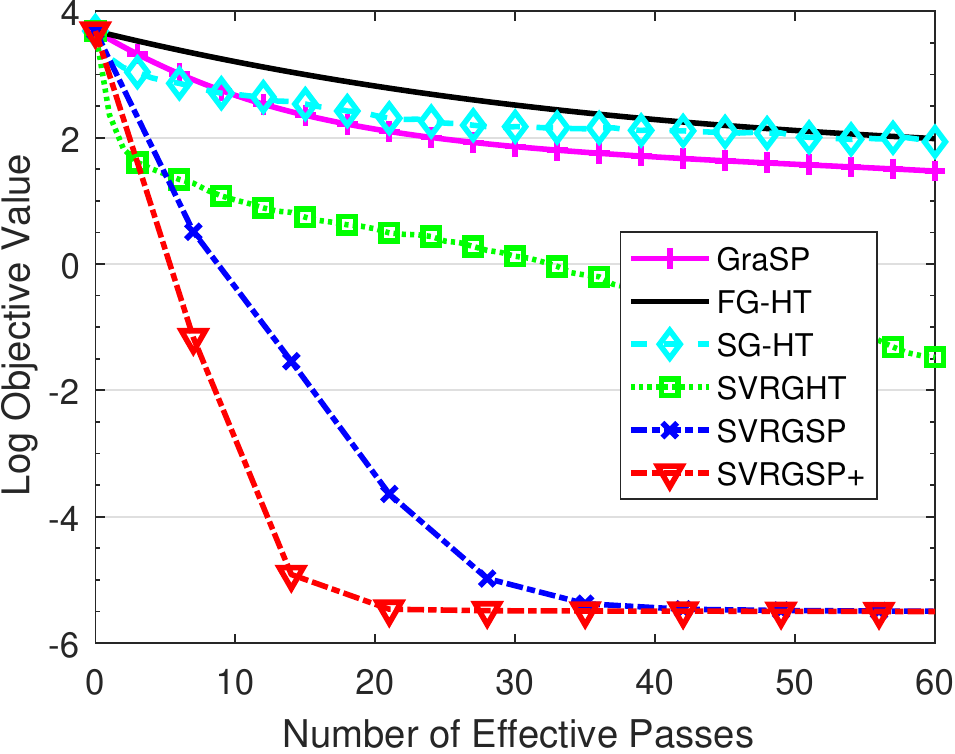}}\;\; \subfigure[$n=5000,\,d=10000,\,s^*\!=500$]{\includegraphics[width=0.505\columnwidth]{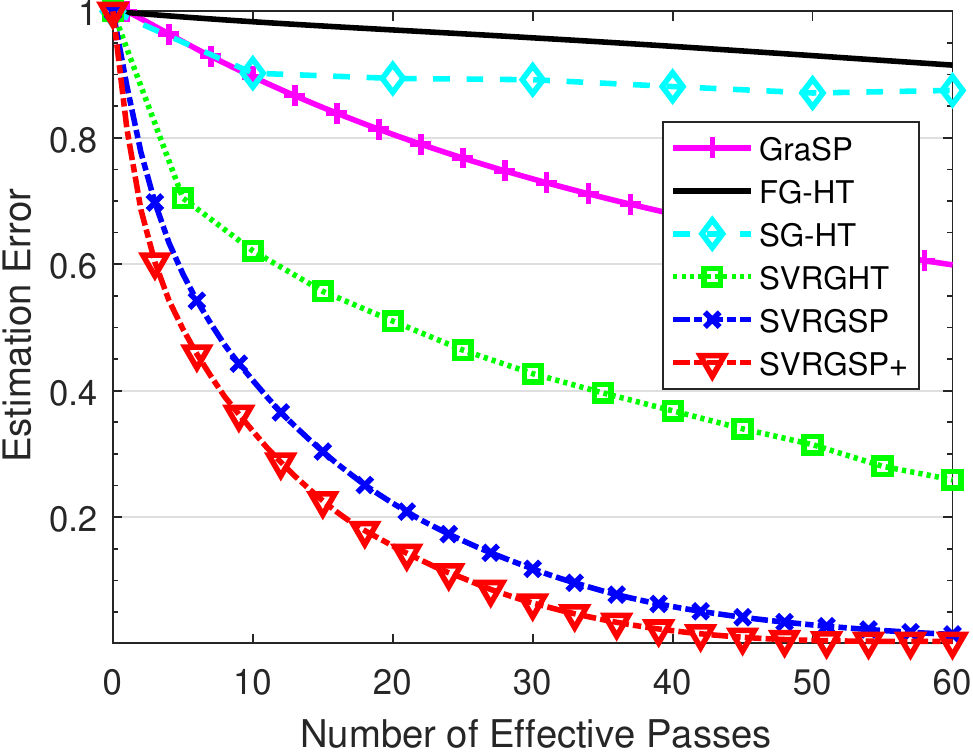}\,\includegraphics[width=0.505\columnwidth]{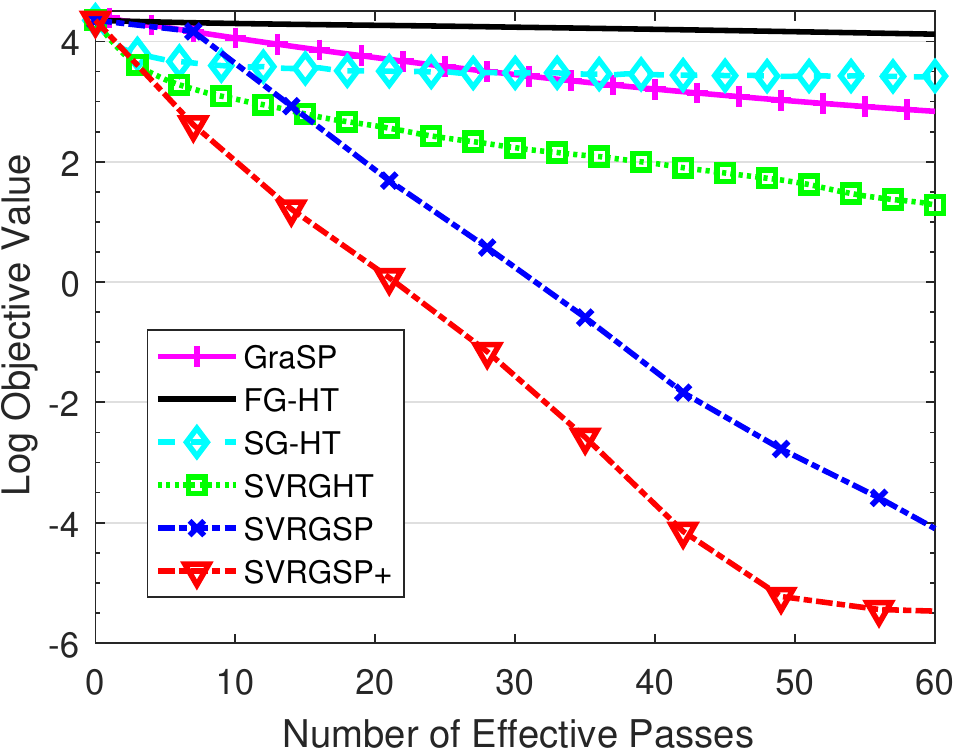}}
\caption{Comparison of all the methods for solving sparsity-constrained linear regression problems on the synthetic datasets. In each plot, the vertical axis denotes the logarithm of the objective function values or estimation error, and the horizontal axis is the number of effective passes over data.}
\label{figs01}
\end{figure*}

\renewcommand{\baselinestretch}{1.39}
\begin{algorithm}[t]
\caption{\!: Our Semi-Stochastic Gradient Solvers}
\label{alg2}
\renewcommand{\algorithmicrequire}{\textbf{Input:}}
\renewcommand{\algorithmicensure}{\textbf{Initialize:}}
\renewcommand{\algorithmicoutput}{\textbf{Output:}}
\begin{algorithmic}[1]
\REQUIRE $\hat{\textbf{x}}^{t-1}$, $|\mathcal{T}|$, and the step-size $\eta$.\\
\ENSURE $m$, $\textbf{g}=\nabla \mathcal{F}(\hat{\textbf{x}}^{t-1})$, $\textbf{z}^{0}=\tilde{\textbf{z}}=\hat{\textbf{x}}^{t-1}$;\\
\FOR{$j=1,2,\cdots,J$}
\STATE {Randomly pick $ i_j \!\in\! \{1,2,\ldots,n\}$;}
\STATE {$\nabla(\textbf{z}^{j-1})=\nabla\! f_{i_j}\!(\textbf{z}^{j-1})-\!\nabla\! f_{i_j}\!(\tilde{\textbf{z}})+\textbf{g}$};
\STATE {$\textbf{z}^{j}\!=\textbf{z}^{j-1}\!-\eta\nabla(\textbf{z}^{j-1})$;~~~// For plain solver\\
\textbf{Option:}~~\textbf{if}~$j$ mod $\lceil J/m\rceil\!=\!0$~\textbf{then}~~~// For fast solver\\
~~~~~~~~~~~~~~~~~~~~~~$\textbf{z}^{j}=\mathcal{H}_{|\mathcal{T}|}(\textbf{z}^{j})$;\\
~~~~~~~~~~~~~~~\textbf{end if}}
\ENDFOR
\OUTPUT {$\textbf{b}=\textbf{z}^{J}$.}
\end{algorithmic}
\end{algorithm}
\renewcommand{\baselinestretch}{1.0}

\subsection{Our Semi-Stochastic Gradient Solvers}
In our RGraSP framework, we apply many semi-stochastic iterations as a solver. Combining our semi-stochastic gradient solver (i.e., Algorithm \ref{alg2}) with our RGraSP framework, we propose a stochastic variance reduced gradient support pursuit (SVRGSP) algorithm and its fast variant (SVRGSP+) to solve Problem (\ref{equ1}). The semi-stochastic gradient solver is outlined in Algorithm \ref{alg2}.

\begin{figure*}[t]
\centering
\subfigure[rcv1]{\includegraphics[width=0.505\columnwidth]{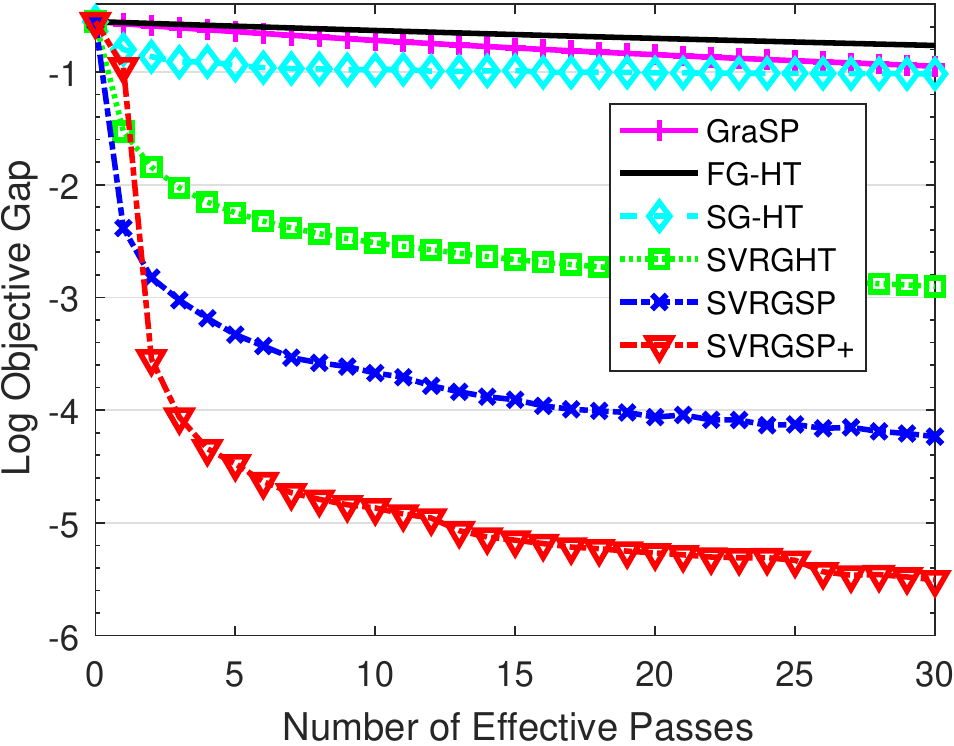}\,\includegraphics[width=0.505\columnwidth]{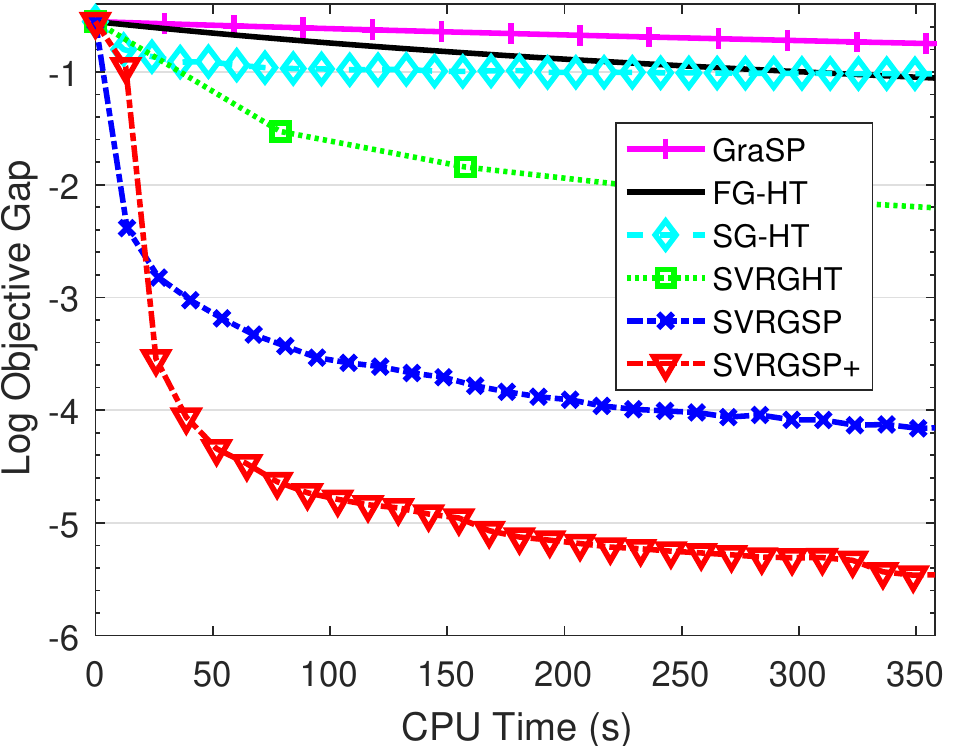}}\;\;
\subfigure[real-sim]{\includegraphics[width=0.505\columnwidth]{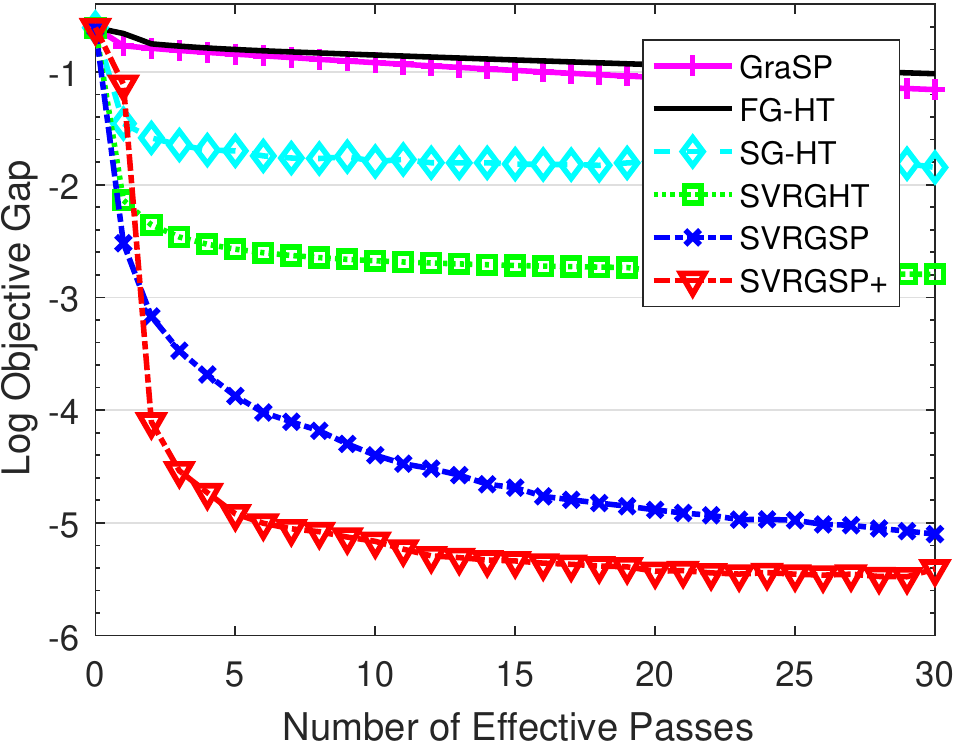}\,\includegraphics[width=0.505\columnwidth]{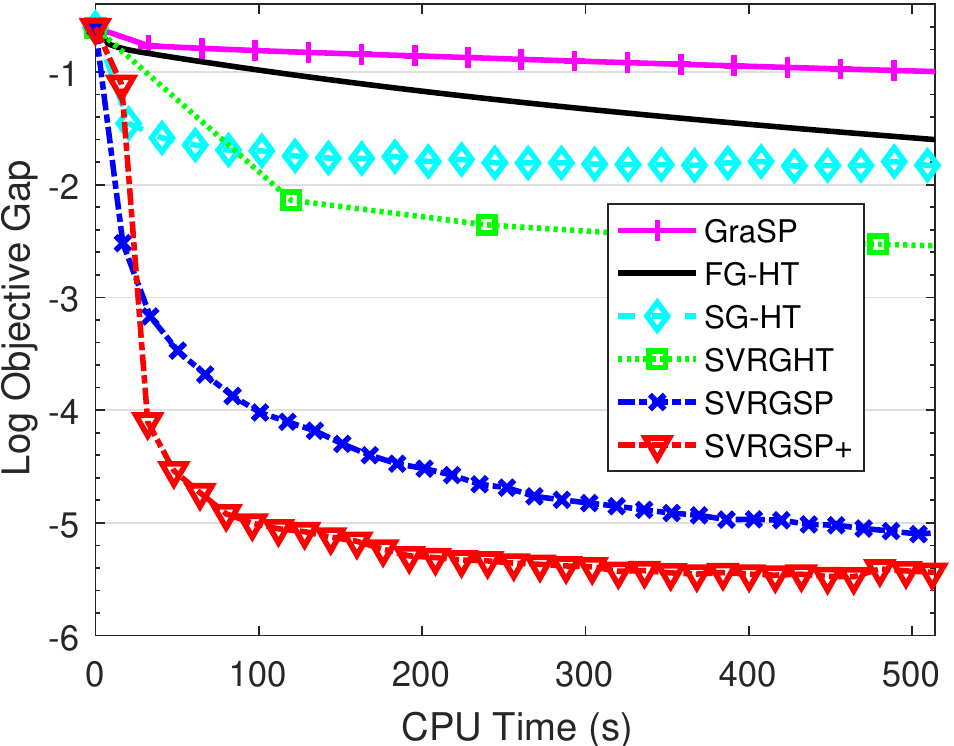}}
\caption{Comparison of all the algorithms for solving sparsity-constrained logistic regression problems on the two real-world datasets. In each plot, the vertical axis is the logarithm of the objective value minus the minimum, and the horizontal axis denotes the number of effective passes over data or running time (seconds).}
\label{figs03}
\end{figure*}

In each iteration of Algorithm \ref{alg2}, we first initialize $\textbf{g}$ as a snapshot gradient in our semi-stochastic gradient update, which has been computed in Algorithm \ref{alg1}. Then we select a sample $i_j$ uniformly at random from $\{1,2,\cdots,n\}$. The semi-stochastic gradient $\nabla(\textbf{z}^{j-1})$ defined in Step 3 of Algorithm \ref{alg2} is updated based on this sample. Note that the gradient $\nabla(\textbf{z}^{j-1})$ is called a semi-stochastic gradient because it includes a deterministic full gradient $\textbf{g}$ and two stochastic gradients, as shown in Step 3 of Algorithm \ref{alg2}. Meanwhile, this semi-stochastic gradient can reduce the variance introduced by randomly sampling and thus can accelerate convergence \cite{DBLP:conf/nips/Johnson013,konevcny2017semi}. Finally, $\textbf{z}^j$ is updated by using the semi-stochastic gradient with a constant step-size $\eta$, as shown in Step 4 of Algorithm \ref{alg2}. Therefore, our SVRGSP solver has no hard thresholding operations in the whole epoch, while existing stochastic algorithms such as SVRGHT \cite{DBLP:journals/corr/LiZALH16} have one hard thresholding operation in each stochastic iteration, which naturally leads to significantly slower convergence.

Moreover, we can also use a few hard thresholding operators (e.g., $m\!=\!6$) in each epoch to maintain the main $|\mathcal{T}|$ coordinates to obtain faster convergence. In other words, Step 4 in Algorithm \ref{alg2} has another option for a fast variant of our SVRGSP solver (i.e., SVRGSP+), which also has significantly less hard thresholding operations than existing algorithms such as SVRGHT \cite{DBLP:journals/corr/LiZALH16}. Therefore, the average per-iteration computational complexity of both SVRGSP and SVRGSP+ is much lower, i.e., $O(d)$ for both SVRGSP and SVRGSP+ vs.\ $O(d\log(d))$ for SVRGHT.

From the above analysis, we can find that our algorithms (i.e., SVRGSP and SVRGSP+) use a hard thresholding operation after a large number of stochastic gradient iterations, while existing stochastic algorithms (e.g., SVRGHT \cite{DBLP:journals/corr/LiZALH16}) perform a hard thresholding operation in each iteration, which is very time-consuming for high-dimensional problems and thus leads to a much slower convergence speed. Although many hard thresholding operations can keep the sparsity of model parameter $\hat{\textbf{x}}^t$, this will lose more gradient information, which is not desirable for fast convergence. In our experiments, we usually set $m\!=\!6$, and $J\!=\!2n$ as the number of iterations. In particular, it is not difficult to prove that our algorithms (i.e., SVRGSP and SVRGSP+) also have a fast linear convergence rate as SVRGHT. Please refer to the long version of this paper for detailed convergence analysis.

\section{Experimental Results}
In this section, we apply the proposed algorithms\footnote{The source codes of our two algorithms can be downloaded from the authors' webpage.} (i.e., SVRGSP and SVRGSP+)  to solve sparsity-constrained linear regression and sparsity-constrained logistic regression problems, and evaluate their empirical performance on many synthetic and real-world datasets. All the experiments were performed on a PC with an Intel i7-7700 CPU and 32GB RAM.

\subsection{Baseline Methods}
In all the experiments, we compare the proposed algorithms (i.e., SVRGSP and SVRGSP+) with the following four state-of-the-art sparsity-constrained algorithms:
\begin{itemize}
  \item Gradient Support Pursuit (GraSP) \cite{DBLP:journals/jmlr/BahmaniRB13};
  \item Fast Gradient descent with Hard Thresholding (FG-HT) \cite{yuan2014gradient};
  \item Stochastic Gradient descent with Hard Thresholding (SG-HT) \cite{DBLP:journals/corr/NguyenNW14};
  \item and Stochastic Variance Reduced Gradient with Hard Thresholding (SVRGHT) \cite{DBLP:journals/corr/LiZALH16}.
\end{itemize}
Note that GraSP and FG-HT are two well-known deterministic optimization methods, while SG-HT and SVRGHT are two recently proposed stochastic optimization methods. The learning rates of all these methods (as well as other parameters) need to be tuned. Here, we set $m\!=\!6$ and $J\!=\!2n$ for our two algorithms. It should be noted that we can get better results by tuning these two parameters together.

\subsection{Synthetic Data}
In this part, we empirically investigate the performance of our algorithms for solving the sparsity-constrained linear model (\ref{equ2}) on many synthetic datasets. We first generate some $n\!\times\! d$ synthetic matrices $\textbf{W}$, each row of which is drawn independently from a $d$-dimensional Gaussian distribution with mean $0$ and covariance matrix $\Sigma \!\in\! \mathbb{R}^{d \times d}$. The response vector is generated from the model $\textbf{y}\!=\!\textbf{W}\textbf{x}^*\!+\!\epsilon$, where $\textbf{x}^*\!\in\! \mathbb{R}^{d}$ is the $s^*$-sparse coefficient vector, and we need to generate the noise $\epsilon$ drawn from a multivariate normal distribution $N (0,\sigma^2 I)$ with $\sigma^2 \!=\!0.01$. The nonzero entries in $\textbf{x}^*$ are sampled independently from a uniform distribution over the interval $[-1,1]$. For the experiments, we construct the following two synthetic data sets:
\begin{enumerate}
  \item $n=2500$, $d=5000$, $s^*\!=250$, $\Sigma\!=\!I$;
  \item $n=5000$, $d=10000$, $s^*\!=500$,
\end{enumerate}
and the diagonal entries of the covariance matrix $\Sigma$ are set to 1, and the other entries are set to $0.1$. The sparsity parameter $s$ is set to $s=1.2 s^*$ for all the algorithms.

Figure \ref{figs01} shows the computational performance (including the logarithm of the objective function values and the estimation error $\frac{\|\hat{\textbf{x}}^{t}-\textbf{x}^*\|_2}{\|\textbf{x}^*\|_2}$) of all the algorithms on the synthetic datasets. All the results show that all the stochastic variance reduction algorithms (i.e., SVRGHT and our SVRGSP and SVRGSP+ algorithms) perform better than the deterministic methods (i.e., GraSP and FG-HT) and the stochastic method, SG-HT. We also can see that both SVRGSP and SVRGSP+ converge significantly faster than all the stare-of-the-art methods in terms of function values and estimation error for all the settings. Although SVRGHT was theoretically proved to have a linear convergence rate for sparsity-constrained linear regression problems, both our algorithms consistently outperform SVRGHT due to less hard thresholding operations, which is consistent with our analysis. Moreover, SVRGSP+ has a slightly faster convergence speed than SVRGSP in all the settings, which validates the importance of a few hard thresholding operators in each epoch.

\subsection{Real-World Data}
We also conduct many experiments on two large-scale real-world datasets: rcv1 and real-sim, which can be downloaded from the LIBSVM Data website\footnote{\url{https://www.csie.ntu.edu.tw/~cjlin/libsvm/}}. They contain $20242$ samples with $47236$ dimensions and $72309$ samples with $20958$ dimensions, respectively, as shown in Table~\ref{sample-table}. We test all the methods for solving sparsity-constrained logistic regression (i.e., $f_i(\textbf{x})\!=\!\log(1\!+\!\exp(-y_i \textbf{w}_i^T \textbf{x}))$) and sparsity-constrained linear regression problems with the sparse parameter $s\!=\!200$ and all the algorithms are initialized with $\hat{\textbf{x}}^{0}\!=\!\textbf{0}$.

\begin{table}
  \caption{Summary of the two large-scale sparse datasets.}
  \label{sample-table}
  \centering
  \setlength{\tabcolsep}{13.5pt}
  \begin{tabular}{llll}
  \toprule
  Datasets     & \#Data     & \#Features  & Sparsity \\
  \midrule
  rcv1 & 20,242 & 47,236  &0.16\%     \\
  real-sim     & 72,309       &20,958   &0.024\%  \\
  \bottomrule
  \end{tabular}
\end{table}

\begin{figure}[t]
\centering
\includegraphics[width=0.497\columnwidth]{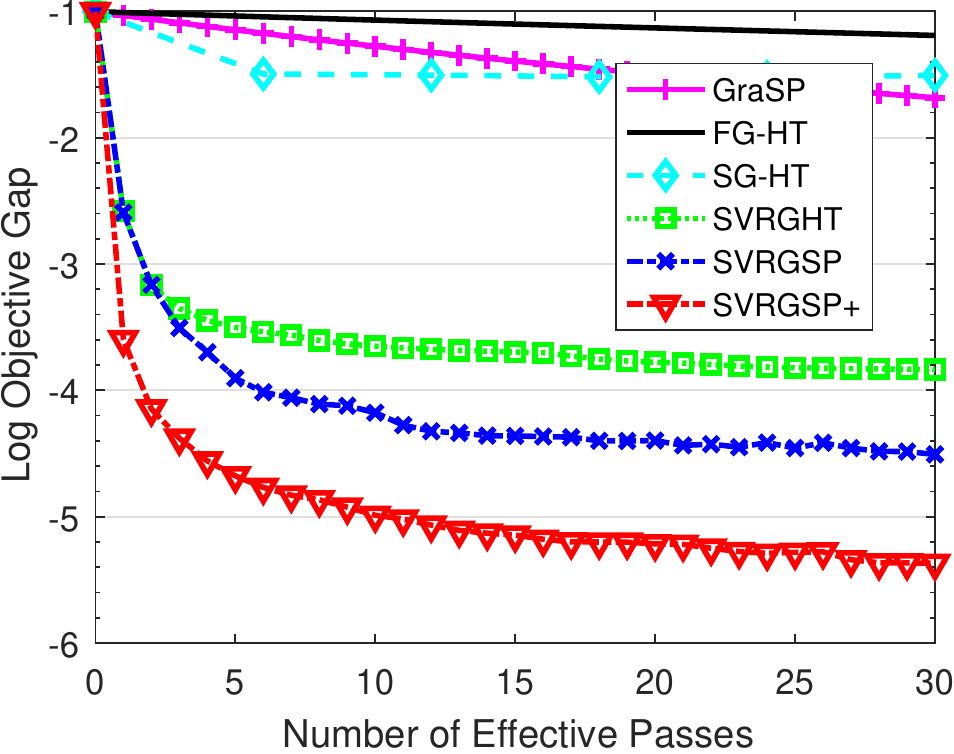}\,\includegraphics[width=0.497\columnwidth]{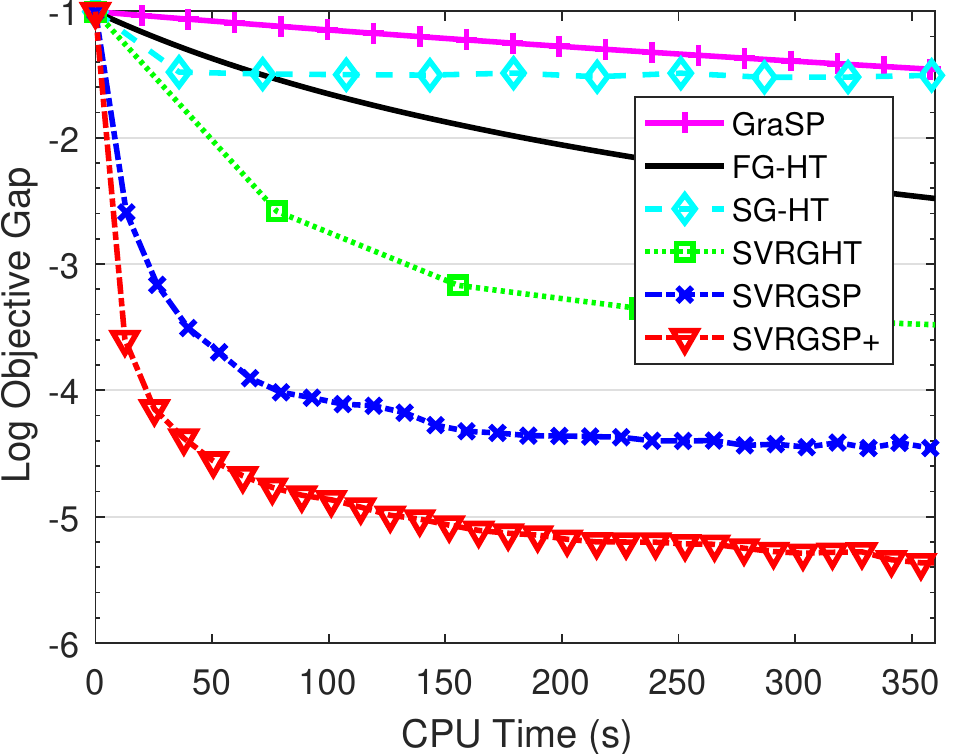}
\caption{Comparison of  all the algorithms for solving sparsity-constrained linear regression problems on rcv1.}
\label{figs04}
\end{figure}

Figure \ref{figs03} shows the logarithm of sparsity-constrained logistic regression function gap (i.e., $\log\|\mathcal{F}(\hat{\textbf{x}}^{t})\!-\!\mathcal{F}(\textbf{x}^*)\|_2$) with respect to the number of effective passes and running time on the rcv1 and real-sim datasets. Similar experimental results of all the algorithms for solving the sparsity-constrained linear regression problem are shown in Figure \ref{figs04}. From all the experimental results, it is clear that the proposed algorithms (i.e., SVRGSP and SVRGSP+) outperform the other state-of-art sparsity-constrained optimization algorithms in terms of both the number of effective passes and running time, meaning that our algorithms (including SVRGSP and SVRGSP+) converge significantly faster than the other algorithms. In particular, SVRGSP+ is consistently faster than all the other algorithms including SVRGSP.

Although the performance of GraSP \cite{DBLP:journals/jmlr/BahmaniRB13} in terms of the number of effective passes is similar to that of FG-HT \cite{yuan2014gradient}, GraSP is usually slower due to its higher per-iteration complexity. Both our algorithms are significantly faster than GraSP in terms of both the number of effective passes and running time, which verifies the effectiveness of our relaxed gradient support pursuit framework. Moreover, all the semi-stochastic descent algorithms (including SVRGHT and our two algorithms) can find some better solutions than the first-order deterministic methods (i.e., GraSP and FG-HT) and the stochastic gradient method, SG-HT, which demonstrates the efficient of the stochastic variance reduced technique. Both our algorithms are much faster than SVRGHT in the all settings, which means that our algorithms are very suitable for real-world large-scale sparse learning problems.

\section{Conclusions and Future Work}
In this paper, we proposed a relaxed gradient support pursuit (RGraSP) framework for solving various large-scale sparsity-constrained optimization problems. Then we also presented two efficient semi-stochastic gradient hard thresholding algorithms as the solver of our RGraSP framework. Our theoretical analysis in the long version of this paper shows that both our algorithms have a fast linear convergence rate. In particular, our algorithms require much less hard thresholding operations than most existing algorithms, and their average per-iteration computational cost is much lower (i.e., $O(d)$ for our algorithms vs.\ $O(d\log(d))$ for the algorithms mentioned above such as SVRGHT), which leads to faster convergence. Various experimental results on synthetic and real-world datasets verified our theoretical results and demonstrated the effectiveness and efficiency of our algorithms.

Unlike GraSP \cite{DBLP:journals/jmlr/BahmaniRB13} and CoSaMP \cite{NeedellT10}, RGraSP is a more general algorithm framework, especially for minimizing many complex cost functions, whose exact solutions cannot be verified in polynomial time. In particular, our proposed semi-stochastic gradient solver is very friendly to asynchronous parallel and distributed implementation similar to~\cite{reddi:sgd,mania:svrg}. Some recently proposed accelerated SVRG algorithms \cite{shang:fsvrg,zhu:Katyusha,zhou:fsvrg,shang:asvrg,shang:gsd,shang:vrsgd} (e.g., Katyusha \cite{zhu:Katyusha} and MiG \cite{zhou:fsvrg}) and their asynchronous parallel and distributed variants can also be used to solve the subproblem in our framework. Moreover, our algorithms and their convergence results can be extended to the non-smooth setting (e.g., non-smooth cost functions) by using the property of stable restricted linearization as in \cite{DBLP:journals/jmlr/BahmaniRB13}, and low-rank matrix and tensor settings such as \cite{shang:lrmf,shang:tsnm,shang:lrtd}.

\section*{Acknowledgments}
This work was supported by the State Key Program of National Natural Science of China (No.\ 61836009), Project supported the Foundation for Innovative Research Groups of the National Natural Science Foundation of China (No.\ 61621005), Major Research Plan of the National Natural Science Foundation of China (Nos.\ 91438201 and 91438103), Fund for Foreign Scholars in University Research and Teaching Programs (the 111 Project) (No.\ B07048), National Natural Science Foundation of China (Nos.\ 61976164, 61876220, 61876221, U1701267, U1730109, and 61871310), Program for Cheung Kong Scholars and Innovative Research Team in University (No.\ IRT\_15R53), Science Foundation of Xidian University (Nos.\ 10251180018 and 10251180019), Fundamental Research Funds for the Central Universities under Grant (No.\ 20101195989), National Science Basic Research Plan in Shaanxi Province of China (No.\ 2019JQ-657), and Key Special Project of China High Resolution Earth Observation System-Young Scholar Innovation Fund.

\bibliographystyle{named}
\balance
\bibliography{ijcai19}

\end{document}